\documentclass[letterpaper, 10 pt, conference]{ieeeconf}  

\usepackage[T1]{fontenc}
\usepackage[utf8]{inputenc}
\usepackage[inkscapearea=page]{svg}
\usepackage[font=small]{caption}
\usepackage{subcaption} 
\usepackage{balance}    
\usepackage{xcolor, soul}  
\usepackage{textpos}    
\usepackage{rotating}   

\usepackage{tikz}
\newcommand*\circled[1]{\tikz[baseline=(char.base)]{
            \node[shape=circle,draw,inner sep=1pt] (char) {#1};}}
            
\usepackage[l2tabu,orthodox]{nag}

\usepackage[
    backend=bibtex8,
    style=ieee,
    sorting=none,
    natbib=true,
    doi=false,
    isbn=false,
    url=false,
    eprint=false,
    maxcitenames=1,
    mincitenames=1
]{biblatex}

\usepackage[pdftex,colorlinks]{hyperref}

\usepackage[printonlyused]{acronym}

\usepackage{siunitx}
\usepackage[all]{nowidow}

\usepackage{lipsum}

\usepackage{xspace} 
\newcommand{\ie}{i.e.,\xspace{}}

\usepackage{epstopdf}

\usepackage{import}

\graphicspath{{./latexGoodPractices/}}

\usepackage{booktabs}

\usepackage{tabularx}
\usepackage{multirow, multicol}

\usepackage{amssymb,amsfonts,amsmath,amscd}

\usepackage{bm}

\newcommand{\bbm}{\begin{bmatrix}}
\newcommand{\ebm}{\end{bmatrix}}

\acrodef{IBC}{Intelligent Boom Control}
\acrodef{IoU}{Intersection over Union}
\acrodef{RPN}{Region Proposal Network}
\acrodef{ROI}{Region-of-Interest}
\acrodef{HTC}{Hybrid Task Cascade}
\acrodef{RRPN}{Rotation Region Proposal Network}
\acrodef{NMS}{Non-Maximum Supression}
\acrodef{CNN}{Convolutional Neural Network}
\acrodef{SGD}{Stochastic Gradient Descent}
\acrodef{LSJ}{large-scale jittering}
\acrodef{mAP}{Mean Average Precision}
\acrodef{AR}{Average Recall}
\acrodef{VIS}{Video Instance Segmentation}

\newcommand\numberofimagesindataset{220 }
\newcommand\numberoflogsindataset{2500 }
\definecolor{purple_1}{rgb}{0.53, 0.0078, 1.0}
\definecolor{red_1}{rgb}{1, 0.4, 0.4}
\definecolor{green_1}{rgb}{0.004, 0.48, 0}

\IEEEoverridecommandlockouts                              

\title{\LARGE \bf
Instance Segmentation for Autonomous Log Grasping in Forestry Operations
}
	\author{Jean-Michel Fortin$^{1}$, Olivier Gamache$^{1}$, Vincent Grondin$^{1}$, François Pomerleau$^{1}$, Philippe Giguère$^{1}$ 
    \thanks{$^{1}$ The authors are with Northern Robotics Laboratory, Université Laval, Québec City, Canada,
    {\tt{\small{$\{$jean-michel.fortin, olivier.gamache, vincent.grondin, francois.pomerleau$\}$ @norlab.ulaval.ca,} philippe.giguere@ift.ulaval.ca}}}%
    \thanks{* This research was supported by the Natural Sciences and Engineering Research Council of Canada (NSERC) through the grant CRD 538321-18, in collaboration with FP Innovations and Resolute Forest Products.}%
    \thanks{** The code and dataset for this paper are available here : \url{https://github.com/norlab-ulaval/logpiles_segmentation}}%
}

\begin{document}

\maketitle
\thispagestyle{empty}
\pagestyle{empty}

\begin{abstract}

Wood logs picking is a challenging task to automate. 
Indeed, logs usually come in cluttered configurations, randomly orientated and overlapping.
Recent work on log picking automation usually assume that the logs' pose is known, with little consideration given to the actual perception problem.  
In this paper, we squarely address the latter, using a data-driven approach.
First, we introduce a novel dataset, named \emph{TimberSeg 1.0}, that is densely annotated, \ie{} that includes both bounding boxes \emph{and} pixel-level mask annotations for logs. 
This dataset comprises \numberofimagesindataset images with 
\numberoflogsindataset individually segmented logs.
Using our dataset, we then compare three neural network architectures on the task of individual logs detection and segmentation; two region-based methods and one attention-based method. 
Unsurprisingly, our results show that axis-aligned proposals, failing to take into account the directional nature of logs, underperform with 19.03 mAP.
A rotation-aware proposal method significantly improve results to 31.83 mAP.
More interestingly, a Transformer-based approach, without any inductive bias on rotations, outperformed the two others, achieving a mAP of 57.53 on our dataset.
Our use case demonstrates the limitations of region-based approaches for cluttered, elongated objects.
It also highlights the potential of attention-based methods on this specific task, as they work directly at the pixel-level.
These encouraging results indicate that such a perception system could be used to assist the operators on the short-term, or to fully automate log picking operations in the future. 

\end{abstract}

\section{INTRODUCTION}


Forestry has seen a large mechanization effort, yet little has been done on automating tasks requiring high-level cognition.
In the last decade, other industries such as agriculture and mining, made significant progress towards automation.
While facing different challenges, forestry is catching up towards autonomous machines in the forest and mills~\citep{oliveira2021advances}. 
Just like the ferrying of ores has been one of the first tasks automated in mining~\citep{marshall2008autonomous}, it is presumed that \emph{forwarding} operations, \ie{} extracting logs from the forest with heavy machinery, will be the first candidate for automation~\cite{ringdahl2011automation}.
Log picking is an essential component in this forwarding task, but is challenging from a perception and manipulation perspective. 
This exacerbates the ongoing manpower shortage in forestry operations, as novice operators require lengthy training to accomplish this repetitive task~\citep{ortiz2014increasing}. 
\begin{figure}[htbp]
	\centering
	\begin{subfigure}[b]{\linewidth}
	    \centering
		\includegraphics[width=\linewidth]{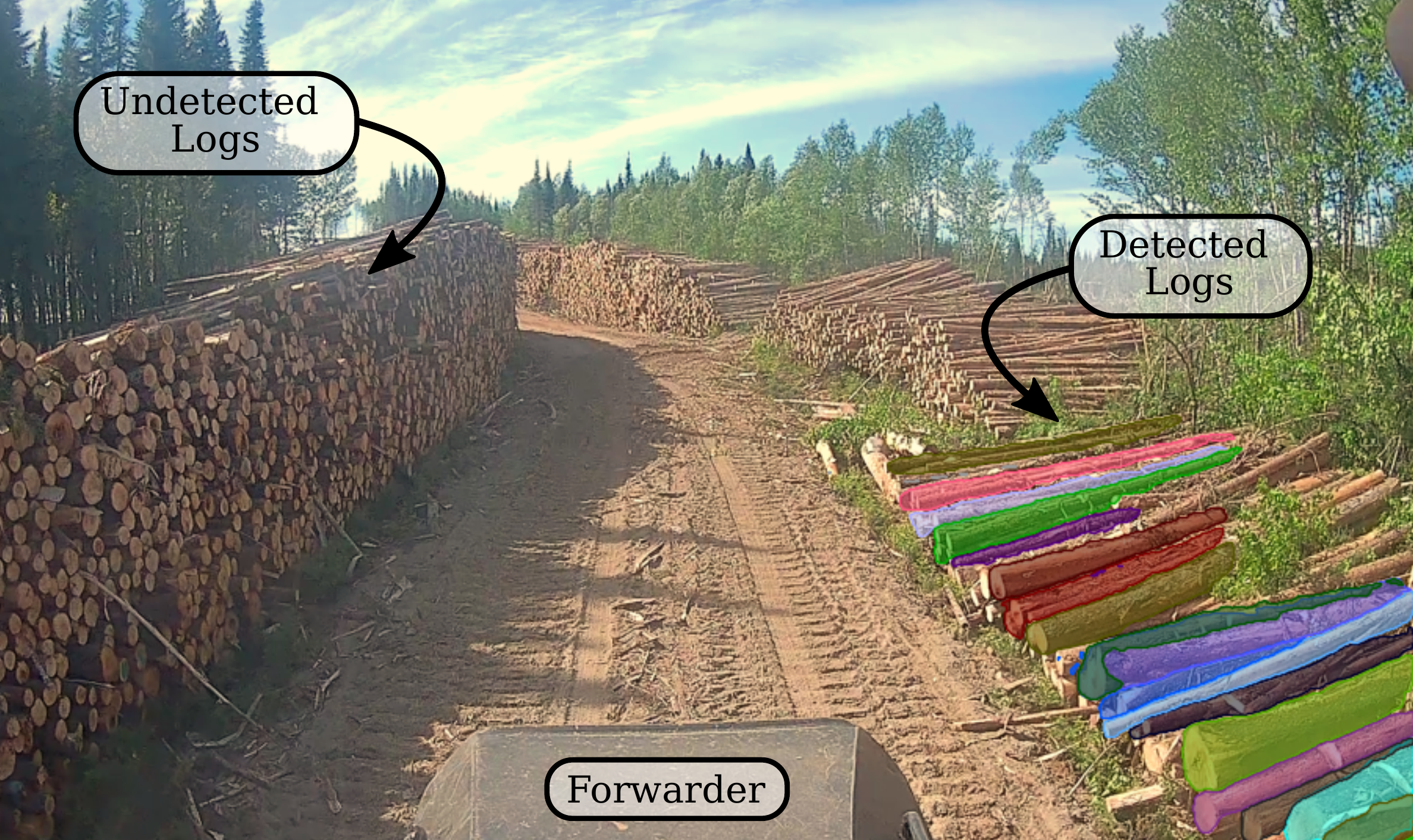}
		\caption{}
		\label{fig:inference_forwarder}
	\end{subfigure}
	\begin{subfigure}[b]{0.45\linewidth}
	    \centering
		\includegraphics[width=\linewidth]{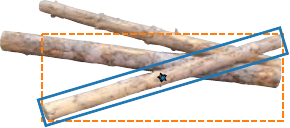}
		\caption{}
		\label{fig:bbox_problem}
	\end{subfigure}
    \hfill
	\begin{subfigure}[b]{0.45\linewidth}
	    \centering
		\includegraphics[width=\linewidth]{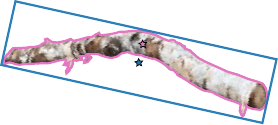}
		\caption{}
		\label{fig:rotated_problem}
	\end{subfigure}
	\caption{\textbf{(a)} View from an actual forestry forwarder, captured with one of our dashcams during actual log picking operations. Mask2Former \citep{cheng2021masked}, the best performing network trained on our dataset, predicts masks on wood logs in the scene (right). We focus on detecting top logs from an upper point of view, hence the undetected logs on the left. \textbf{(b)} Rotated bounding boxes (solid blue line) are more adapted to detect logs because the latter are elongated and randomly oriented. 
	This is in contrast with axis-aligned bounding boxes (dashed orange line), which will encompass other logs.
	\textbf{(c)} For localization purposes, oriented boxes are not enough since crooked logs lead to the center point being off, as shown by the stars. Therefore, we focused on instance segmentation with masks.}
	\label{fig:motivation}
\end{figure}
On the short run, teleoperation~\cite{persson2020design} could allow operators to work from remote locations, while having access to better situational awareness with real-time navigation and augmented reality. 
Other assistance systems, like the \ac{IBC}, are also simplifying the work of operators, thus reducing their cognitive load~\cite{zemanek2022influence}. 
Taking advantage of these technological advances, unmanned logging machines are getting closer to reality. 
However, previous work on autonomous log handling~\citep{andersson2021reinforcement} assume that the position and orientation of the logs are known, which is generally not the case. 

Our work focuses primarily on this log perception problem, targeting autonomous forestry forwarders and wood loaders.
In computer vision, earlier work on object detection focused on predicting bounding boxes encompassing the target objects. 
The object's center of mass, the usual grasp point for a log, could then be assumed to be in the middle of the bounding box.
However, this assumption will fail for non-rectangular objects such as crooked wood logs, as seen in \autoref{fig:rotated_problem}.
Thus, our system aims at segmenting individual logs in RGB images from a camera placed in or above the operator's cabin, as shown in \autoref{fig:inference_forwarder}. 
One reason that detection of wood logs is non-trivial is that they tend to be cluttered by nature. 
Indeed, logs are elongated objects that are generally found either randomly spread on the ground, or in piles throughout the recently harvested forest or the lumberyard. Moreover, they suffer from partial occlusion, either from branches or simply other logs, which makes instance segmentation more challenging.
Variations in illumination and weather in outdoor locations are also challenging, especially in northern countries such as Canada or Sweden.
In these areas, forestry operations are performed throughout the winter, thus in snowy conditions. 




Our main contribution consists in characterizing data-driven approaches, namely deep learning, on the task of wood logs segmentation. 
In this regard, we provide a \emph{densely annotated} dataset with a diversity of scenarios, \ie{} piled and individual logs, various seasons, environments, illuminations and point of views. 
We also demonstrate the viability of speeding up manual data labeling by using an earlier-trained network to suggest labels to a human annotator, simplifying future annotation jobs.
We then tested three instance segmentation architectures. 
First, we used Mask R-CNN~\cite{he2017mask}, a well-known and widely used segmentation network, as a baseline. Its poor results show that using axis-aligned bounding boxes as region proposals leads to sub-par quality masks, mainly because they encompass other logs, as depicted in \autoref{fig:bbox_problem}. 
Since logs are elongated and not axis-aligned, an upgraded version using oriented anchors~\cite{ma2018arbitrary} in the \ac{RPN} was considered. 
This helps by reducing the presence of other logs in the \ac{ROI}. 
Finally, we tested a recent, state-of-the-art transformer network called Mask2Former~\cite{cheng2021masked} which shows promising results on the log segmentation task.
In particular, we show that this type of approach seems to be compatible with long-shaped, cluttered objects. 

In short, the contributions of this work are: 
\begin{itemize}
    \item a demonstration of the potential and applicability of different deep neural networks architectures on the challenging task of log detection; and
    \item \emph{TimberSeg 1.0}: the first publicly available dataset of \numberofimagesindataset RGB images of logs in various environments, with dense annotation (bounding boxes and masks) for \numberoflogsindataset logs.
\end{itemize}

\section{RELATED WORK}

Our survey of the literature focuses on recent instance segmentation methods and their variants, targeting applicability towards detecting elongated, cluttered and overlapping objects. We also analyze some recent applications of deep learning in forestry and agriculture, highlighting the absence of work and datasets in autonomous log picking. 


\subsection{Instance Segmentation Architectures}

Segmentation is usually performed by adding an object mask prediction to existing object detection methods for bounding box prediction. 
We can classify instance segmentation approaches in multiple sub-categories below.

\textbf{Single-stage methods} 
work by directly making predictions from the feature maps extracted by the backbone, often in a grid-like manner.
One example is YOLACT++~\cite{bolya2020yolact++}, which is extending RetinaNet's~\cite{lin2017focal} dense grid prediction by adding a mask branch that combines prototype masks extracted from the feature maps. 
The SOLOv2 network~\cite{wang2020solov2} instead deals directly with instance segmentation, where each square of a grid is responsible for generating a segmentation mask and a class label, without generating bounding boxes.

\textbf{Two-stage methods} are named so because of the intermediary step of predicting \ac{ROI}s, followed by a second stage classifying and refining the localization of the object. 
Faster R-CNN~\cite{ren2015faster} was the first to use a \acf{RPN} to predict these \ac{ROI}s, which are refined with a regression network.
Mask R-CNN~\cite{he2017mask} later added a mask branch for instance segmentation.
Whereas single-stage methods are more computationally efficient, two-stage methods yield better accuracy and recall on the widely used COCO dataset~\citep{lin2014microsoft}.




\textbf{Rotated proposal} methods were subsequently developed to target randomly oriented objects with elongated geometries. 
Indeed, Mask R-CNN displays strong results on datasets containing regular-shaped objects such as COCO~\cite{lin2014microsoft}.
However, the axis-aligned and rectangular-shaped assumption of its \ac{RPN} is challenged when trained on cluttered or randomly oriented long objects, such as scattered wood logs. 
Thus,~\citet{ma2018arbitrary} proposed a novel \ac{RRPN}, in their case for detecting non-axis aligned text.
They added an angle factor to the anchor boxes, bounding objects more precisely, as shown in \autoref{fig:bbox_problem}.
This improved Rotated Mask R-CNN method was successfully applied
on remote sensing detection of ships in~\cite{koo2018rbox}, which exhibit some visual similarities with scattered logs. 

\textbf{Attention-based methods}~\cite{vaswani2017attention} are radically departing from previous approaches, putting forward universal frameworks for all segmentation tasks. It started with DETR~\citep{carion2020end}, using a transformer encoder-decoder and queries to detect a fixed number of objects. 
QueryInst~\citep{fang2021instances} further adapted DETR to instance segmentation, by adding dynamic mask heads. 
Also based on DETR, MaskFormer~\citep{cheng2021per} first segments the input image into a fixed number of masks and then classifies them.
Incremental changes  brought forth by Mask2Former~\citep{cheng2021masked} provided faster convergence and higher accuracy. 
Because Mask2Former works directly at the pixel level through attention, they no longer include the inductive bias of rectangular regions.
Consequently, they can achieve better accuracy on instance segmentation of logs (\ie{} cluttered, elongated objects), as we will demonstrate later.

\vspace{10mm}

\subsection{Deep Learning in Forestry and Agriculture}

Many papers are applying vision and deep learning
in forestry and agriculture. 
Specifically, the problem of instance segmentation of fruits and vegetables has received significant attention.
For example,~\citet{halstead2020fruit} applied Faster R-CNN and Mask R-CNN for sweet pepper detection over multiple domains, demonstrating the natural ability of the model to generalize when provided diverse enough data. 
Mask R-CNN was also evaluated for strawberries detection~\citep{yu2019fruit} in a fruit-picking robot. 
Their system visually separated ripe from unripe fruits, using the predicted masks for stem localization. 
~\citet{liu2019cucumber} slightly modified Mask R-CNN for cucumber fruits detection, using the assumption that cucumbers are always green.
They used color filtering to remove background pixels, thus accelerating the generation of anchor boxes in \ac{RPN}.
Their results show that masks are necessary for precise center point estimation of elongated, non-linear objects such as cucumbers, which relates well to wood logs. 
However, we show in our experiments that Mask R-CNN is hardly applicable when objects are superimposed and close to each other, partly because of the \ac{NMS} step.
Moreover, the lack of uniformity in logs' color, shape and texture, as well as distinctiveness with the background precludes any color or texture filtering approach.

In forestry, we notice a growing interest in public datasets for deep learning tasks such as classification and detection. \citet{carpentier2018tree} released a large bark images dataset for tree species identification. 
\citet{da2021visible} published a dataset of standing trees for bounding boxes detection and trained a YOLOv3~\cite{redmon2018yolov3} network on it. 
~\citet{li2021implementation} trained this same network for detecting ground obstacles in a harvested forest, on a custom dataset.
Nevertheless, the latter two datasets only provide bounding boxes ground truths, which can be useful to visually assist the machine operator, but hardly applicable to precise localization. 
Moreover, they rely heavily on the axis-aligned assumption, which may be suitable for standing trees, but not for randomly piled wood logs.

The problem of detecting and localizing logs, both in harvested forests or in lumberyards, has received little attention. 
More than a decade ago, \citet{park20113d} demonstrated the potential of a structured light-based vision system to detect and localize logs, predating deep learning. 
Experiments were conducted indoor, greatly diminishing its applicability to outdoor environments. 
More recently, ~\citet{polewski2021instance} used convolutional neural networks for semantic segmentation of fallen trees in aerial images, separating instances using shape parametrization.
This technique is unsuitable for log piles, since the masked pile can be blob-shaped, making it impossible to distinguish individual instances geometrically.
More related to forwarders and wood loaders, motion planning to grasp logs is analyzed in~\citep{andersson2021reinforcement}.
There, they used reinforcement learning to control the crane motion of a forwarder.
However, the target log's pose is assumed to be known in the aforementioned work. 
Our paper thus seeks to fill the gap in the literature around wood logs detection, while providing data to push research forward.
Indeed, we conjecture that one of the reasons that explains the lack of research on wood logs detection is the absence of publicly available labeled datasets. 


 \begin{figure}[tbp]
	\centering
	\begin{subfigure}[b]{0.46\linewidth}
	    \centering
		\includegraphics[width=\linewidth]{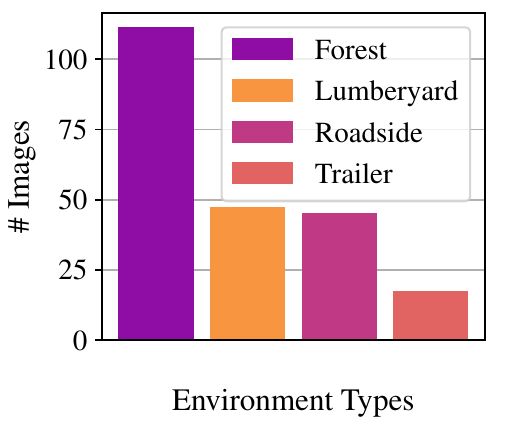}
		\caption{}
		\label{fig:environment_types}
	\end{subfigure}
	~
	\begin{subfigure}[b]{0.46\linewidth}
	    \centering
		\includegraphics[width=\linewidth]{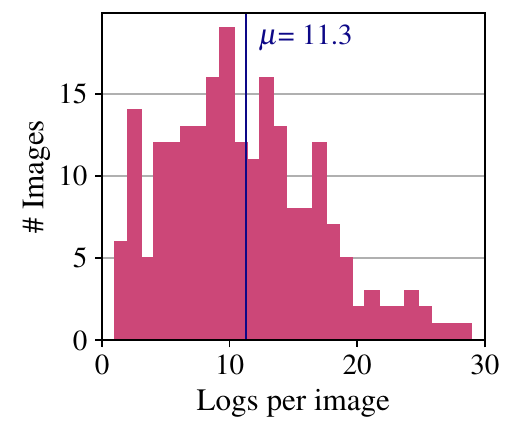}
		\caption{}
		\label{fig:logs_per_image}
	\end{subfigure}
	
    ~
	\begin{subfigure}[b]{0.46\linewidth}
	    \centering
		\includegraphics[width=\linewidth]{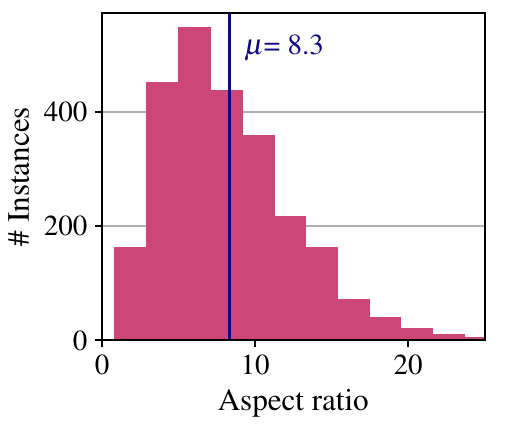}
		\caption{}
		\label{fig:aspect_ratio}
	\end{subfigure}
	~
	\begin{subfigure}[b]{0.46\linewidth}
	    \centering
		\includegraphics[width=\linewidth]{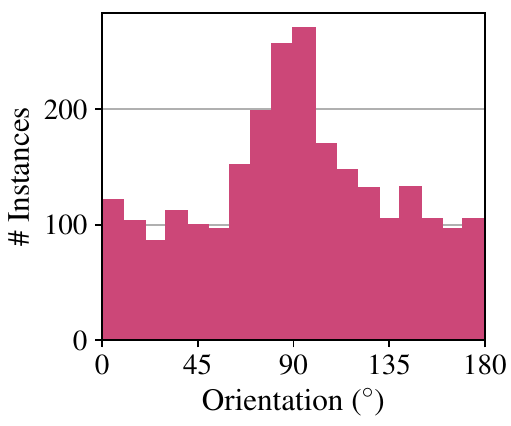}
		\caption{}
		\label{fig:orientation}
	\end{subfigure}
	\caption{Dataset statistics. Top graphs show the variety of images in the dataset, with images taken in different strategic locations with the number of visible logs per image ranging from 1 to 30. Bottom graphs demonstrate how logs are elongated and randomly oriented, making axis-aligned bounding boxes irrelevant.}
	\label{fig:histogram}
\end{figure}

\section{METHODOLOGY}

In this section, we first detail our approach for data collection and labeling. Then, we describe each evaluated network architecture, to further highlight their specificities in regards to our task at hand. Lastly, we provide all the training details.

\begin{figure*}[htbp]
	\centering
	\includegraphics[width=\linewidth]{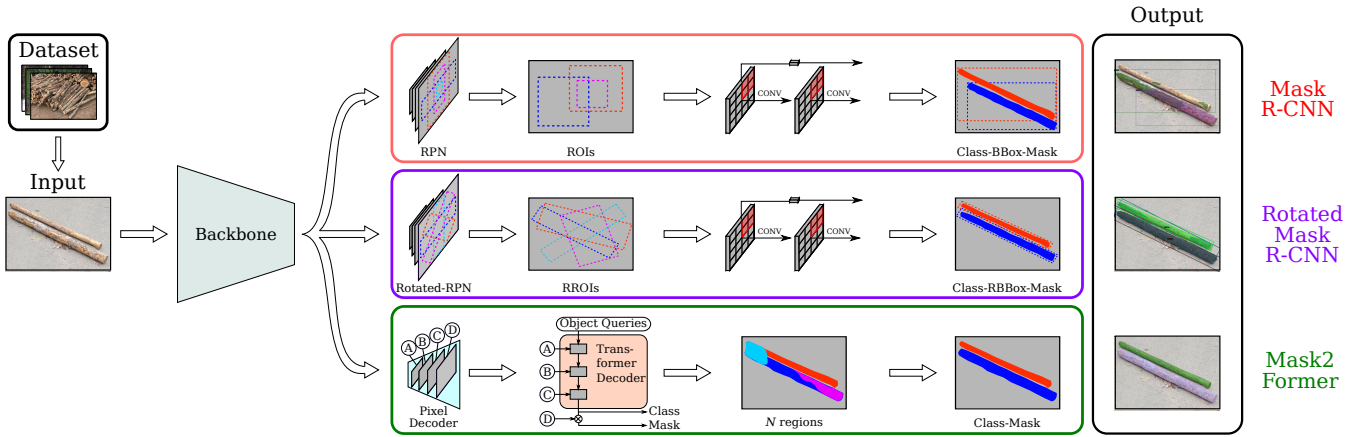}
	\caption{High-level comparison between tested instance segmentation networks. Each of the three networks takes as input an image from our dataset, first passing through a backbone for features extraction. \textcolor{red_1}{\textbf{Mask R-CNN}} determines \ac{ROI}s using an \ac{RPN} with axis-aligned anchor boxes. The network's heads then extract bounding boxes and masks from these \ac{ROI}s. \textcolor{purple_1}{\textbf{Rotated Mask R-CNN}} uses rotated anchor boxes to select rotated \ac{ROI}s and uses similar heads to refine them into oriented bounding boxes and masks. \textcolor{green_1}{\textbf{Mask2Former}} does not use region proposals, instead working with multi-scale deformable attention on the whole feature maps, at the pixel-level. The transformer decoder uses self-attention to segment \textit{N} regions from \textit{N} trainable object queries and we filter the predicted masks by score. The output for each network is displayed in the right column, for an actual image from our dataset.}
	\label{fig:comparison_networks}
\end{figure*}

\subsection{Dataset Collection and Labeling}
\label{sec:labeling}

The dataset used in our experiments, called \emph{TimberSeg 1.0}, is composed of \numberofimagesindataset images containing \numberoflogsindataset instances of wood logs in representative environments and dispositions.
Thus, these images were carefully selected to capture typical viewpoints and situations 
that a forwarder or wood loader operator would operate with.
We can see in \autoref{fig:environment_types} that images have been collected in four types of environments. 
\emph{Forest} corresponds mostly to scattered logs in freshly cut areas, whereas \emph{roadside} consists of neatly piled logs ready to be transported, as seen in \autoref{fig:inference_forwarder}. 
To collect this data, three dashcams (\emph{VIOFO A129 Pro Duo 4K}) were installed in forestry forwarders operating in forests near Lake Saint-Jean, Quebec, Canada.
Each dashcam comprised two cameras, fixed on the inner side of the cabin's windows.
One camera pointed forward, while the other pointed sideways.
They recorded hundreds of hours of video, at various resolutions (4K, 2K, 1080p) over multiple months and weather conditions. 
Pictures, using a \emph{Canon EOS M50} digital camera, were also taken in \emph{lumberyards} from sawmills and paper mills, including images of filled timber trucks' \emph{trailers} from an upper viewpoint. 
The combination of dynamic images coming from the videos and static images taken with the camera provides additional diversity to the dataset. 
Finally, complementary images were taken from publicly available videos from the Internet, to further increase the geographical diversity \cite{halstead2020fruit}. 
The forests images are predominant on purpose, as they exhibit harder visual conditions.
The images contain a widely varying number of log instances as depicted in \autoref{fig:logs_per_image}, ranging from 1 to 29 logs, for an average of 11.3.
\autoref{fig:aspect_ratio} shows the aspect ratio of logs, which resembles a log-normal distribution, with a mean value of 8.3 and a tail distribution going as high as a ratio of 24.
This is far from the 2:1 or 1:2 aspect ratios typically used for anchor boxes, reflecting the non-standardness of our problem.
The distribution of log orientations, presented in \autoref{fig:orientation}, shows a peak around $90^{\circ}$, indicating that a significant portion of the logs are parallel to the $y$-axis of the camera.
Otherwise, logs do not display any other dominant orientation in images, making their detection challenging to axis-aligned proposal approaches.

	

Our dataset was annotated using polygon mask segmentation with the \emph{SuperAnnotate} tool. 
From a log picking perspective, it is sufficient to detect and segment only the logs on top of a pile.
Others will become accessible as the pile is collected. 
In this regard, a log is only segmented if it is accessible to the forwarder, meaning that a log covered by others or too far away should not be labeled.
Similarly to active learning approaches, we use our most accurate network (Mask2Former) to generate pre-annotations and progressively accelerate the labeling process. 
A human annotator then corrects the mistakes made by the network instead of segmenting everything from scratch.


\subsection{Network Architectures}


Three instance segmentation architectures are evaluated and compared on our dataset: a vanilla Mask R-CNN, a Rotated Mask R-CNN and a Mask2Former network. \autoref{fig:comparison_networks} shows a visual depiction of the key elements of each architecture.

\subsubsection{Mask R-CNN}
Given the historical importance of Mask R-CNN~\cite{he2017mask}, it will serve as a baseline for region-proposal methods. 
First, a \ac{CNN} backbone processes the input image to produce multi-resolution feature maps.
Then, the \ac{RPN} uses a sliding window to evaluate the objectness of each predefined axis-aligned anchor box, to generate region proposals. 
The network's heads independently predict a class, a bounding box and a pixel-wise mask for every \ac{ROI}. 
Testing Mask R-CNN will help highlight the limitations of such axis-aligned approaches for our task at hand.

\subsubsection{Rotated Mask R-CNN}
~\citet{ma2018arbitrary} showed that Mask R-CNN has limitations when objects in the scene are dense or arbitrarily oriented. 
As a potential solution, they introduce a rotation-based framework and propose the \acf{RRPN}, which adds an orientation to anchor boxes. 
This improvement makes \ac{ROI}s fit more tightly around objects (see \autoref{fig:bbox_problem}), reducing the amount of irrelevant pixels and reaching for more accurate masks.

\subsubsection{Mask2Former}

Attention-based architectures like Mask2Former~\citep{cheng2021masked} work directly at the pixel-level. 
Similarly to Mask R-CNN, a backbone first extracts a pyramid of feature maps.
The usual backbone for this method is a Swin Transformer~\citep{liu2021swin}, but a \ac{CNN} backbone would also work. 
A six-layer multi-scale deformable attention Transformer (MSDeformAttn)~\citep{zhu2020deformable} then decodes the feature maps to generate high-resolution per-pixel embeddings.
Finally, a transformer decoder uses masked attention to process these embeddings as well as object queries to output masks and labels. 
This decoder is composed of nine layers handling different resolutions of feature maps, depicted in a simplified manner in \autoref{fig:comparison_networks} by the \circled{A}-\circled{D} labels.

\subsection{Training Details}
We used the {\tt{Detectron2}}\footnote{\tt{\url{https://github.com/facebookresearch/detectron2}}} library based on PyTorch \citep{paszke2019pytorch} to train the models on single-class detection. 
Our Mask R-CNN implementation mostly follows~\citep{he2017mask}, with a ResNeXt-101 backbone. 
For training, the initial learning rate was set to 0.001, reducing by a factor of 5 every 2000 iterations, with mini-batches of size 8. 
The model was optimized using an \ac{SGD} method with 0.9 momentum and 0.0001 weight decay. 
We trained for 8000 iterations, with the test set containing \SI{20}{\%} on the images.
The initial network's weights had been previously pretrained on ImageNet-1k, for the backbone, and then on COCO dataset for the whole network.

The only changes applied to the rotated version of Mask R-CNN are the use of \ac{RRPN} and RRoIAlign~\citep{ma2018arbitrary} for oriented region proposals. 
Box anchors were selected to have a 30 degrees orientation interval and aspect ratios of 4, 8, 12 and 16, following the dataset statistics displayed in \autoref{fig:aspect_ratio}. Both \ac{RPN} and \ac{RRPN} use a smooth-L1 loss over the box parameters ($x$, $y$, width, height and angle) where ($x$,$y$) are the center coordinates. 
We increased the number of detections per image to 250 instead of the original 100, which slightly improves accuracy and recall.

Mask2Former was also trained following most recommendations in the original paper~\cite{cheng2021masked}. 
The feature extractor is a Swin-B backbone with a window size of 12, pretrained on ImageNet-1k and COCO.
We set the number of object queries for the decoder to 100. 
We used the same learning schedule and hyperparameters as Mask R-CNN, except for the use of AdamW optimizer and a weight decay of 0.02.

We use the same data augmentation for each network, following the \ac{LSJ} method proposed in~\citet{ghiasi2021simple}, which was proven to be highly data efficient. 
It consists of randomly resizing the images in a scale between 0.1 and 2.0 as well as cropping to a fixed size of 1024 $\times$ 1024 pixels with gray padding and flipping horizontally.
We also normalized images with ImageNet's mean and standard deviation for each color channel.    
The hardware used for model training and testing consists of a single computer with a \emph{Nvidia RTX-3090} GPU, an \emph{Intel Core i9-10900KF} CPU and 64 GB of RAM. Single fold training time for Mask R-CNN was 2.5 hours, while Rotated Mask R-CNN took 5 hours per fold and Mask2Former only 45 minutes.

\section{EXPERIMENTS \& RESULTS}

Multiple experiments were conducted to compare and analyze the performances of selected networks on \emph{TimberSeg 1.0}. 
In this section, we start by comparing each architecture, both quantitatively and qualitatively.
We seek to corroborate our previous hypothesis that region-based approaches hinder good instance segmentation, in the case of cluttered and elongated objects. 
We then present further experiments on the best performing network, Mask2Former, to better characterize its performances. 
All results are generated using a 5-fold cross-validation policy, to fully leverage our dataset. 

\subsection{Networks Comparison}

We evaluated the predicted masks using the standard COCO metrics \ac{mAP} and AP50, as well as \ac{AR} and the F1-score, each time keeping the 100 best scoring detections. 
\autoref{tab:comparison_networks} presents a quantitative comparison between the three architectures on the task of instance segmentation of wood logs.
We first note an important improvement of over \SI{12}{\%} in \ac{mAP} when switching to rotated region proposals instead of using axis-aligned boxes.
More importantly, it shows that Mask2Former has a strong edge over two-stage networks on our application, nearly doubling all metrics compared to the rotated version of Mask R-CNN.
Pretraining the Mask2Former backbone on ImageNet-22k yields only marginal improvements.

\autoref{tab:comparison_networks} also displays the inference speed for each network. The lower frame rate obtained by Rotated Mask R-CNN can be explained by the higher number of anchor box proposals. 
Indeed, a standard \ac{RPN} uses 15 anchor box shapes, whereas our \ac{RRPN} implementation employs 84 because of the multiple orientations. 
Despite using less computation than \ac{RRPN}, Mask2Former exhibits a significant accuracy improvement. 
Overall, it shows that detection in real time or near real time is possible, opening the door to visual-servoing methods. 

\begin{table*}[tbp]
    \def\arraystretch{1.2}
    \setlength{\tabcolsep}{5.5pt}
    \centering
    \caption{Performance comparison of each tested network on our dataset, evaluated on predicted masks. Results show that using rotated regions in Mask R-CNN significantly improves accuracy, recall and thus F1-score. Mask2Former outperforms both approaches, using attention mechanisms based on Transformers \cite{vaswani2017attention}. Pretraining the Swin-B backbone on ImageNet-22k further improves results by a small margin.}
    \begin{tabular}{@{}ll|cccc|c@{}}
         \hline
         Architecture & Backbone & mAP & AP50 & Recall & F1-score & fps \\ 
         \hline \hline
         Mask R-CNN & X101-FPN & $19.03_{\pm3.21}$ & $36.10_{\pm4.45}$ & $28.60_{\pm3.98}$ & $0.23_{\pm0.04}$ & 12.55 \\ \hline
         Rot. Mask R-CNN & X101-FPN & $31.83_{\pm3.26}$ & $52.78_{\pm4.33}$ & $36.95_{\pm3.14}$ & $0.34_{\pm0.03}$ & 5.66 \\ \hline
         Mask2Former & Swin-B (1k) & $56.05_{\pm3.12}$ & $82.97_{\pm1.99}$ & $64.06_{\pm3.05}$ & $0.60_{\pm0.03}$ & 8.47 \\ \hline
         Mask2Former & Swin-B (22k) & $\textbf{57.53}_{\pm3.37}$ & $\textbf{84.28}_{\pm2.44}$ & $\textbf{65.16}_{\pm3.40}$ & $\textbf{0.61}_{\pm0.03}$ & 8.47 \\ \hline
    \end{tabular}
    
    \label{tab:comparison_networks}
\end{table*}

\begin{figure}[bp]
	\centering
	\includegraphics[width=\linewidth]{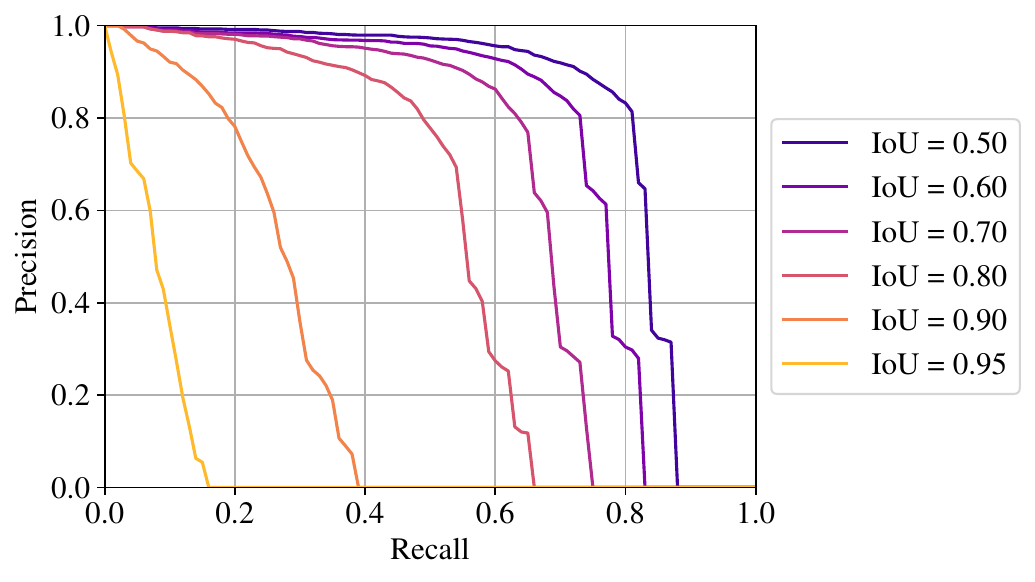}
	\caption{Precision-Recall curves for various mask \ac{IoU}s using our best performing Mask2Former network. Model shows impressive performances for \ac{IoU}s up to 0.7.}
	\label{fig:pr_curve}
\end{figure}


\autoref{fig:pr_curve} shows the precision-recall curves for our Mask2Former implementation, evaluated using mask \ac{IoU}. As in \autoref{tab:comparison_networks}, these results are for single-frame detection.
When taking the 0.5 \ac{IoU} curve, we can see that a precision greater than \SI{80}{\%} can be achieved while maintaining an \SI{80}{\%} recall. In the context of log picking, a low recall implies that some logs would be left behind, while a low precision means that the grapple often misses the target. Since humans should not be around during these operations, missing the target would not have devastating consequences other than slowing down operations. 

A better metric for log grasping would be the success rate, tested in real-life conditions. 
In the absence of a machine to evaluate such a metric, we considered that an \ac{IoU} of 0.5 should be sufficient for a successful grasp, yielding a precision of \SI{84.39}{\%} as displayed in \autoref{tab:comparison_networks}.
This takes into account the fact that the center of mass of the predicted mask would be in the actual log.
Moreover, the opening width of a grapple is several times the diameter of a log, accommodating for ample localization errors.
Finally, although an autonomous forwarder would need to automatically choose the next target log, it is not vital to pick the optimal one and, following our labeling protocol (\autoref{sec:labeling}), the predicted logs should always be the ones on top of the pile and accessible.

\subsection{Power Law and Real-Time Potential} 
\begin{figure}[htbp]
	\centering
	\begin{subfigure}[b]{0.48\linewidth}
	    \centering
		\includegraphics[width=\linewidth]{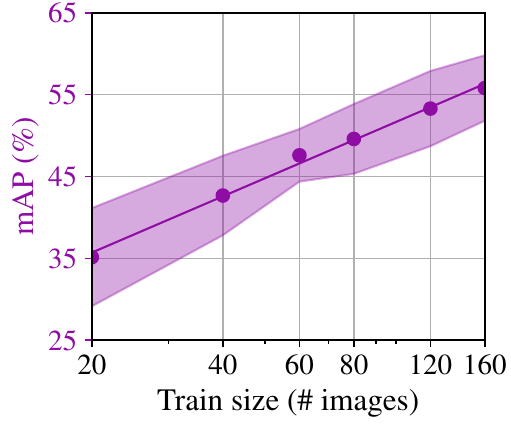}
		\caption{}
		\label{fig:ablation_train}
	\end{subfigure}
	~
	\begin{subfigure}[b]{0.48\linewidth}
	    \centering
		\includegraphics[width=\linewidth]{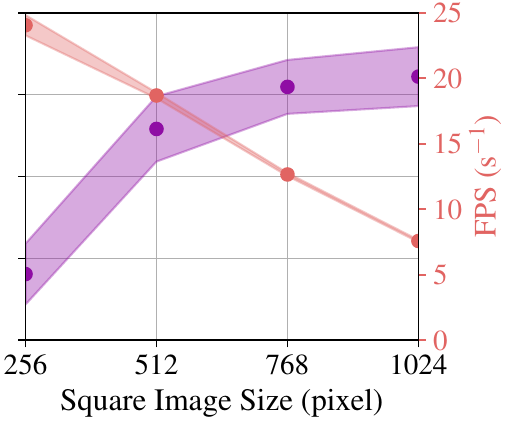}
		\caption{}
		\label{fig:ablation_image}
	\end{subfigure}
	\caption{Performances analysis of Mask2Former network.
	Bands are showing one standard deviation.
	(a) The \ac{mAP} follows a power law, as a function of the number training images.
	Each doubling of the training dataset size yields an improvement of \SI{6.87}{\%} on average.
	(b) Image resolution impacts both inference speed and accuracy.
	Increasing the input image size in Mask2Former's network leads to better performances, but limits real-time applications.
	}
	\label{fig:ablation}
\end{figure}

A fundamental question when trying to tackle a problem using data-driven approaches is to estimate the amount of labeled data needed to solve it.
To this effect, we investigated the relationship between this training dataset size and the observed \ac{mAP} with Mask2Former.
There, the number of training images ranged from 20 to 160, with a fixed test set of 44 images.
A linear regression on the semi-log plot, as depicted by the purple line in \autoref{fig:ablation_train}, shows that we are in the presence of a power law, as seen also by others using Vision Transformers~\cite{Zhai2021ScalingVT}.
With some reserves, we can expect an improvement of around \SI{6.87}{\%} each time the dataset size is doubled, until a plateau is reached.
In this regard, it is safe to assume that leveraging a bigger dataset or semi-supervised methods would lead to better predictions.
These are, however, beyond the scope of this work.

In \autoref{fig:ablation_image}, we ran another experiment, this time varying the size of the image at the network input.
As the resolution increases, the network's accuracy improves, in line with what is observed in the literature.
Conversely, frame rate decreases due to increased computation needs to process larger images.
By varying the input image size, one can get an interesting trade-off between speed and accuracy. For instance, the frame rate can go as high as \SI{24.05}{fps} when the image size is reduced to 256 $\times$ 256, accepting that precision will drop accordingly. We observe an interesting sweet spot around an image size of 512 $\times$ 512, where the inference speed can be tripled with only a \SI{5}{\%} drop in precision.

\subsection{Qualitative Results} 
\begin{figure*}[htbp]
    \begin{textblock}{0.5}(0.03,0.25)
        \begin{turn}{90}\small Ground Truth\end{turn}
    \end{textblock}
    \begin{textblock}{0.5}(0.03,1.8)
        \begin{turn}{90}\small Predictions\end{turn}
    \end{textblock}
    \begin{textblock}{0.5}(0.03,3.55)
        \begin{turn}{90}\small Ground Truth\end{turn}
    \end{textblock}
    \begin{textblock}{0.5}(0.03,5.1)
        \begin{turn}{90}\small Predictions\end{turn}
    \end{textblock}
	\centering
	\begin{subfigure}[b]{0.48\linewidth}
	    \centering
		\includegraphics[width=\linewidth]{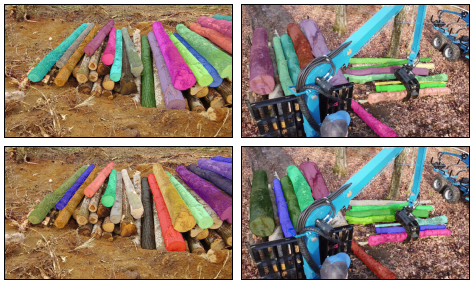}
		\caption{Clean}
		\label{fig:autumn}
	\end{subfigure}
	\hspace{-2.3mm}
	\begin{subfigure}[b]{0.48\linewidth}
	    \centering
		\includegraphics[width=\linewidth]{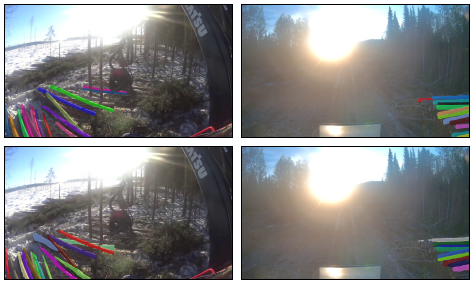}
		\caption{Sun glare}
		\label{fig:sun_glare}
	\end{subfigure}
    \\
    \vspace{1.5mm}
	\begin{subfigure}[b]{0.48\linewidth}
	    \centering
		\includegraphics[width=\linewidth]{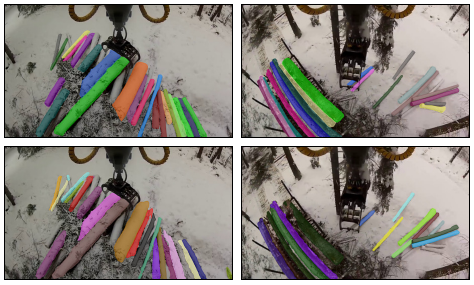}
		\caption{Winter}
		\label{fig:snow}
	\end{subfigure}
	\hspace{-2.3mm}
	\begin{subfigure}[b]{0.48\linewidth}
	    \centering
		\includegraphics[width=\linewidth]{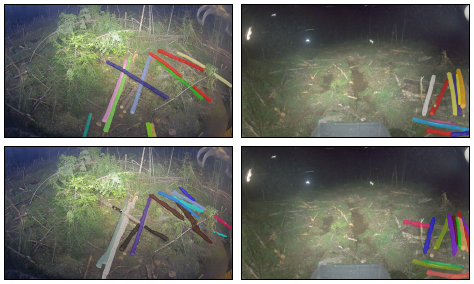}
		\caption{Night}
		\label{fig:night}
	\end{subfigure}
	\caption{Examples of segmentation results from trained Mask2Former network on our dataset (bottom row) compared to ground truth labels (top row). Notice how the network is capable of obtaining good results in the presence of distractors such as sunglare~(b), winter conditions~(c) and vignetting due to artificial illumination during night operation~(d).}
	\label{fig:comparison_qualitative}
\end{figure*}

\autoref{fig:comparison_qualitative} shows a number of examples of images from our dataset, with their annotation, for difficult images.
Prediction outputs from Mask2Former are also provided in this figure. 
The network displays good robustness to challenging conditions such as snow, sun glare and night time operation.

\section{CONCLUSION}
In this paper, we addressed the problem of log picking automation in forestry operations.
To this effect, we created a novel dataset of log images, some of them taken during the actual operation of a forwarder.
The \emph{TimberSeg 1.0} dataset, which will be made public upon publication, contains \numberofimagesindataset images with \numberoflogsindataset logs precisely annotated with bounding boxes and masks.
Using our dataset, we then compared the applicability of three deep neural network architectures on the task of wood log segmentation. 
Our results showed that Mask2Former, an approach based on attention, works best on this type of cluttered, occluded and randomly oriented objects, as opposed to the more traditional region-proposal based methods. 
Our results demonstrate the viability of such a perception system, which could be used for operators’ aid in teleoperation or as a stepping stone towards autonomous forwarders and wood loaders.
Our validation also opens the door to automatic roadside logpile inventories, by simply using a camera mounted on a vehicle.

For future works, the success rate yielded by our approach needs to be validated on a real wood loader the ensure the maturity of the solution. Also, detection accuracy could be improved in several ways.
On the data side, we plan on leveraging the large volume of unlabeled videos that we collected with our dashcams. 
Semi- and self-supervised learning methods, such as Masked Auto-encoders~\cite{MAE2021}, could then be used in pretraining the backbone to our application domain.
Improved data augmentation, such as \emph{copy-paste}~\cite{ghiasi2021simple} could also be employed to reduce dependency on labeled images.
On the algorithm side, our approach is currently single frame.
As such, our results constitute a worst-case scenario, where planning of log grasps would be based on detections coming from a single image.
Techniques based on Bayesian filtering could be employed to improve detection reliability, explicitly tracking log instances across video frames.
Implicit approaches performing detection directly from video streams \citep{yang2019video} could also be envisioned.
\balance






\printbibliography

@article{ortiz2014increasing,
  title={Increasing the {L}evel of {A}utomation in the {F}orestry {L}ogging {P}rocess with {C}rane {T}rajectory {P}lanning and {C}ontrol},
  author={Ortiz Morales, Daniel and Westerberg, Simon and La Hera, Pedro X and Mettin, Uwe and Freidovich, Leonid and Shiriaev, Anton S},
  journal={{J}ournal of {F}ield {R}obotics},
  volume={31},
  number={3},
  pages={343--363},
  year={2014},
  publisher={Wiley Online Library}
}

@article{oliveira2021advances,
  title     = {{A}dvances in {F}orest {R}obotics: {A} {S}tate-of-the-{A}rt {S}urvey},
  author    = {Oliveira, Luiz FP and Moreira, Ant{\'o}nio P and Silva, Manuel F},
  journal   = {Robotics},
  volume    = {10},
  number    = {2},
  pages     = {53},
  year      = {2021},
  publisher = {Multidisciplinary Digital Publishing Institute}
}

@article{li2021implementation,
  title     = {{Implementation of a System for Real-Time Detection and Localization of Terrain Objects on Harvested Forest Land}},
  author    = {Li, Songyu and Lideskog, H{\aa}kan},
  journal   = {Forests},
  volume    = {12},
  number    = {9},
  pages     = {1142},
  year      = {2021},
  publisher = {Multidisciplinary Digital Publishing Institute}
}

@article{da2021visible,
  title     = {{Visible and Thermal Image-Based Trunk Detection with Deep Learning for Forestry Mobile Robotics}},
  author    = {da Silva, Daniel Queir{\'o}s and Dos Santos, Filipe Neves and Sousa, Armando Jorge and Filipe, V{\'\i}tor},
  journal   = {Journal of Imaging},
  volume    = {7},
  number    = {9},
  pages     = {176},
  year      = {2021},
  publisher = {Multidisciplinary Digital Publishing Institute}
}

@inproceedings{carpentier2018tree,
  title        = {{Tree Species Identification from Bark Images Using Convolutional Neural Networks}},
  author       = {Carpentier, Mathieu and Gigu\`ere, Philippe and Gaudreault, Jonathan},
  booktitle    = {IEEE/RSJ International Conference on Intelligent Robots and Systems},
  pages        = {1075--1081},
  year         = {2018}
}

@mastersthesis{persson2020design,
  title  = {Design of a workstation for teleoperated forwarders: {E}xploring the future work within forestry},
  author = {Persson, Tobias},
  school = {Luleå University of Technology},
  year   = {2020}
}

@phdthesis{ringdahl2011automation,
  title  = {Automation in forestry: development of unmanned forwarders},
  author = {Ringdahl, Ola},
  year   = {2011},
  school = {Institutionen f{\"o}r datavetenskap, Ume{\aa} Universitet}
}

@article{andersson2021reinforcement,
  title   = {{R}einforcement {L}earning {C}ontrol of a {F}orestry {C}rane {M}anipulator},
  author  = {Andersson, Jennifer and Bodin, Kenneth and Lindmark, Daniel and Servin, Martin and Wallin, Erik},
  journal = {arXiv:2103.02315},
  year    = {2021}
}

@inproceedings{park20113d,
  title        = {{3D Log Recognition and Pose Estimation for Robotic Forestry Machine}},
  author       = {Park, Yeonchool and Shiriaev, Anton and Westerberg, Simon and Lee, Sukhan},
  booktitle    = {IEEE International Conference on Robotics and Automation},
  pages        = {5323--5328},
  year         = {2011}
}

@article{polewski2021instance,
  author   = {Polewski, Przemyslaw and Shelton, Jacquelyn and Yao, Wei and Heurich, Marco},
  doi      = {https://doi.org/10.1016/j.isprsjprs.2021.06.016},
  issn     = {0924-2716},
  journal  = {ISPRS Journal of Photogrammetry and Remote Sensing},
  pages    = {297--313},
  title    = {{Instance segmentation of fallen trees in aerial color infrared imagery using active multi-contour evolution with fully convolutional network-based intensity priors}},
  url      = {https://www.sciencedirect.com/science/article/pii/S092427162100174X},
  volume   = {178},
  year     = {2021}
}

@inproceedings{halstead2020fruit,
  title     = {{Fruit Detection in the Wild: The Impact of Varying Conditions and Cultivar}},
  author    = {Halstead, Michael and Denman, Simon and Fookes, Clinton and McCool, Chris},
  booktitle = {IEEE {D}igital {I}mage {C}omputing: {T}echniques and {A}pplications},
  pages     = {1--8},
  year      = {2020}
}

@article{liu2019cucumber,
  title     = {{Cucumber Fruits Detection in Greenhouses Based on Instance Segmentation}},
  author    = {Liu, Xiaoyang and Zhao, Dean and Jia, Weikuan and Ji, Wei and Ruan, Chengzhi and Sun, Yueping},
  journal   = {IEEE {A}ccess},
  volume    = {7},
  pages     = {139635--139642},
  year      = {2019},
  publisher = {IEEE}
}

@article{yu2019fruit,
  title     = {{Fruit detection for strawberry harvesting robot in non-structural environment based on Mask-RCNN}},
  author    = {Yu, Yang and Zhang, Kailiang and Yang, Li and Zhang, Dongxing},
  journal   = {Computers and {E}lectronics in {A}griculture},
  volume    = {163},
  pages     = {104846},
  year      = {2019},
  publisher = {Elsevier}
}

@article{ma2018arbitrary,
  title     = {{Arbitrary-Oriented Scene Text Detection via Rotation Proposals}},
  author    = {Ma, Jianqi and Shao, Weiyuan and Ye, Hao and Wang, Li and Wang, Hong and Zheng, Yingbin and Xue, Xiangyang},
  journal   = {IEEE Transactions on Multimedia},
  volume    = {20},
  number    = {11},
  pages     = {3111--3122},
  year      = {2018},
  publisher = {IEEE}
}

@inproceedings{koo2018rbox,
  title     = {{RBox-CNN: Rotated Bounding Box based CNN for Ship Detection in Remote Sensing Image}},
  author    = {Koo, Jamyoung and Seo, Junghoon and Jeon, Seunghyun and Choe, Jeongyeol and Jeon, Taegyun},
  booktitle = {ACM International Conference on Advances in Geographic Information Systems (SIGSPATIAL)},
  pages     = {420--423},
  year      = {2018}
}

@inproceedings{ghiasi2021simple,
  title     = {Simple {C}opy-{P}aste is a {S}trong {D}ata {A}ugmentation {M}ethod for {I}nstance {S}egmentation},
  author    = {Ghiasi, Golnaz and Cui, Yin and Srinivas, Aravind and Qian, Rui and Lin, Tsung-Yi and Cubuk, Ekin D and Le, Quoc V and Zoph, Barret},
  booktitle = {IEEE/CVF Conference on Computer Vision and Pattern Recognition},
  pages     = {2918--2928},
  year      = {2021}
}

@inproceedings{ren2015faster,
  title   = {{Faster R-CNN: Towards Real-Time Object Detection with Region Proposal Networks}},
  author  = {Ren, Shaoqing and He, Kaiming and Girshick, Ross and Sun, Jian},
  booktitle = {Advances in Neural Information Processing Systems},
  volume  = {28},
  pages   = {91--99},
  year    = {2015}
}

@inproceedings{he2017mask,
  title     = {Mask {R-CNN}},
  author    = {He, Kaiming and Gkioxari, Georgia and Doll{\'a}r, Piotr and Girshick, Ross},
  booktitle = {IEEE International Conference on Computer Vision},
  pages     = {2961--2969},
  year      = {2017}
}

@article{redmon2018yolov3,
  title   = {{YOLOv3: An Incremental Improvement}},
  author  = {Redmon, Joseph and Farhadi, Ali},
  journal = {arXiv preprint arXiv:1804.02767},
  year    = {2018}
}

@article{bolya2020yolact++,
  title     = {{YOLACT++ Better Real-Time Instance Segmentation}},
  author    = {Bolya, Daniel and Zhou, Chong and Xiao, Fanyi and Lee, Yong Jae},
  journal   = {IEEE Transactions on Pattern Analysis and Machine Intelligence},
  volume = {44},
  number = {2},
  pages = {1108-1121},
  year      = {2022}
}

@inproceedings{lin2017focal,
  title     = {{Focal Loss for Dense Object Detection}},
  author    = {Lin, Tsung-Yi and Goyal, Priya and Girshick, Ross and He, Kaiming and Doll{\'a}r, Piotr},
  booktitle = {IEEE International Conference on Computer Vision},
  pages     = {2980--2988},
  year      = {2017}
}

@article{wang2020solov2,
  title   = {{SOLOv2: Dynamic and Fast Instance Segmentation}},
  author  = {Wang, Xinlong and Zhang, Rufeng and Kong, Tao and Li, Lei and Shen, Chunhua},
  journal = {Advances in Neural Information Processing Systems},
  volume  = {33},
  pages   = {17721--17732},
  year    = {2020}
}

@inproceedings{vaswani2017attention,
  title   = {{Attention Is All You Need}},
  author  = {Vaswani, Ashish and Shazeer, Noam and Parmar, Niki and Uszkoreit, Jakob and Jones, Llion and Gomez, Aidan N and Kaiser, {\L}ukasz and Polosukhin, Illia},
  booktitle = {Advances in Neural Information Processing Systems},
  volume  = {30},
  pages   = {6000--6010},
  year    = {2017}
}

@inproceedings{carion2020end,
  title        = {{End-to-End Object Detection with Transformers}},
  author       = {Carion, Nicolas and Massa, Francisco and Synnaeve, Gabriel and Usunier, Nicolas and Kirillov, Alexander and Zagoruyko, Sergey},
  booktitle    = {European Conference on Computer Vision},
  pages        = {213--229},
  year         = {2020}
}

@inproceedings{cheng2021per,
  title   = {{Per-Pixel Classification is Not All You Need for Semantic Segmentation}},
  author  = {Cheng, Bowen and Schwing, Alex and Kirillov, Alexander},
  booktitle = {Advances in Neural Information Processing Systems},
  volume  = {34},
  year    = {2021}
}

@article{cheng2021masked,
  title   = {{M}asked-attention {M}ask {T}ransformer for {U}niversal {I}mage {S}egmentation},
  author  = {Cheng, Bowen and Misra, Ishan and Schwing, Alexander G and Kirillov, Alexander and Girdhar, Rohit},
  journal = {arXiv preprint arXiv:2112.01527},
  year    = {2021}
}

@inproceedings{liu2021swin,
  title     = {Swin {T}ransformer: {H}ierarchical {V}ision {T}ransformer using {S}hifted {W}indows},
  author    = {Liu, Ze and Lin, Yutong and Cao, Yue and Hu, Han and Wei, Yixuan and Zhang, Zheng and Lin, Stephen and Guo, Baining},
  booktitle = {IEEE/CVF International Conference on Computer Vision},
  pages     = {10012--10022},
  year      = {2021}
}

@inproceedings{fang2021instances,
  title     = {{Instances as Queries}},
  author    = {Fang, Yuxin and Yang, Shusheng and Wang, Xinggang and Li, Yu and Fang, Chen and Shan, Ying and Feng, Bin and Liu, Wenyu},
  booktitle = {IEEE/CVF International Conference on Computer Vision},
  pages     = {6910--6919},
  year      = {2021}
}

@article{zhu2020deformable,
  title   = {Deformable {DETR}: {D}eformable {T}ransformers for {E}nd-to-{E}nd {O}bject {D}etection},
  author  = {Zhu, Xizhou and Su, Weijie and Lu, Lewei and Li, Bin and Wang, Xiaogang and Dai, Jifeng},
  journal = {arXiv preprint arXiv:2010.04159},
  year    = {2020}
}

@article{zemanek2022influence,
  title     = {{I}nfluence of {I}ntelligent {B}oom {C}ontrol in {F}orwarders on {P}erformance of {O}perators},
  author    = {Zem{\'a}nek, Tom{\'a}{\v{s}} and Fil'o, Petr},
  journal   = {Croatian Journal of Forest Engineering},
  volume    = {43},
  number    = {1},
  pages     = {47--64},
  year      = {2022},
  publisher = {Fakultet {\v{s}}umarstva i drvne tehnologije Sveu{\v{c}}ili{\v{s}}ta u Zagrebu}
}

@article{marshall2008autonomous,
  title={Autonomous underground tramming for center-articulated vehicles},
  author={Marshall, Joshua and Barfoot, Timothy and Larsson, Johan},
  journal={{J}ournal of {F}ield {R}obotics},
  volume={25},
  number={6-7},
  pages={400--421},
  year={2008},
  publisher={Wiley Online Library}
}

@inproceedings{lin2014microsoft,
  title={{Microsoft COCO: Common Objects in Context}},
  author={Lin, Tsung-Yi and Maire, Michael and Belongie, Serge and Hays, James and Perona, Pietro and Ramanan, Deva and Doll{\'a}r, Piotr and Zitnick, C Lawrence},
  booktitle={European Conference on Computer Vision},
  pages={740--755},
  year={2014}
}

@inproceedings{yang2019video,
  title={Video {I}nstance {S}egmentation},
  author={Yang, Linjie and Fan, Yuchen and Xu, Ning},
  booktitle={IEEE/CVF International Conference on Computer Vision},
  pages={5188--5197},
  year={2019}
}

@article{MAE2021,
  author    = {Kaiming He and
               Xinlei Chen and
               Saining Xie and
               Yanghao Li and
               Piotr Doll{\'{a}}r and
               Ross B. Girshick},
  title     = {Masked Autoencoders Are Scalable Vision Learners},
  journal   = {ArXiv},
  volume    = {abs/2111.06377},
  year      = {2021}
}

@article{Zhai2021ScalingVT,
  title={Scaling {V}ision {T}ransformers},
  author={Xiaohua Zhai and Alexander Kolesnikov and Neil Houlsby and Lucas Beyer},
  journal={ArXiv},
  year={2021},
  volume={abs/2106.04560}
}

@article{paszke2019pytorch,
  title={Pytorch: An imperative style, high-performance deep learning library},
  author={Paszke, Adam and Gross, Sam and Massa, Francisco and Lerer, Adam and Bradbury, James and Chanan, Gregory and Killeen, Trevor and Lin, Zeming and Gimelshein, Natalia and Antiga, Luca and others},
  journal={Advances in neural information processing systems},
  volume={32},
  year={2019}
}

\end{document}